\documentclass[sigconf,screen]{acmart}

\AtBeginDocument{
  }

\acmSubmissionID{4395}

\usepackage{booktabs}
\usepackage{amsmath} 
\usepackage{graphicx} 
\usepackage{multirow}
\usepackage{makecell}
\usepackage{algorithm}
\usepackage{algpseudocode}
\usepackage{subcaption}
\begin{document}

\title[DA-PTQ: Drift-Aware Post-Training Quantization]{DA-PTQ: Drift-Aware Post-Training Quantization \\ for Efficient Vision-Language-Action Models}

\author{Siyuan Xu}
\affiliation{
  \institution{Tongji University}
  \city{Shanghai}
  \country{China}
}
\email{2352539@tongji.edu.cn}

\author{Tianshi Wang}
\affiliation{
  \institution{Tongji University}
  \city{Shanghai}
  \country{China}
}
\email{tswang0116@163.com}

\author{Fengling Li}
\affiliation{%
  \institution{University of Technology Sydney}
  \city{Sydney}
  \country{Australia}
}
\email{fenglingli2023@gmail.com}

\author{Lei Zhu}
\affiliation{%
  \institution{Tongji University}
  \city{Shanghai}
  \country{China}
}
\email{leizhu0608@gmail.com}

\author{Heng Tao Shen}
\affiliation{%
  \institution{Tongji University}
  \city{Shanghai}
  \country{China}
}
\email{shenhengtao@hotmail.com}

\renewcommand{\shortauthors}{Xu et al.}

\begin{abstract}
Vision-Language-Action models (VLAs) have demonstrated strong potential for embodied AI, yet their deployment on resource-limited robots remains challenging due to high memory and computational demands. While Post-Training Quantization (PTQ) provides an efficient solution, directly applying PTQ to VLAs often results in severe performance degradation during sequential control. We identify temporal error accumulation as a key factor, where quantization perturbations at the vision-language-to-action interface are progressively amplified, leading to kinematic drift in executed trajectories. To address this issue, we propose Drift-Aware Post-Training Quantization (DA-PTQ), which formulates quantization as a drift-aware optimization problem over sequential decision processes. DA-PTQ consists of two components: (1) Cross-Space Representation Compensation, which mitigates structured distortions between multimodal representations and action space to improve action consistency, and (2) Motion-Driven Mixed-Precision Allocation, which assigns bit-widths by minimizing trajectory-level motion errors. Extensive experiments show that DA-PTQ significantly reduces kinematic drift and achieves comparable performance to full-precision models under low-bit settings, enabling practical deployment of VLAs on resource-limited robotic platforms.
\end{abstract}

\ccsdesc[500]{Computing methodologies~Artificial intelligence}
\ccsdesc[500]{Computing methodologies~Vision for robotics}

\keywords{Vision-Language-Action Models, Post-Training Quantization, Kinematic Drift, Efficient Reasoning}

\maketitle

\begin{figure}[htbp]
    \centering
    \includegraphics[width=\linewidth]{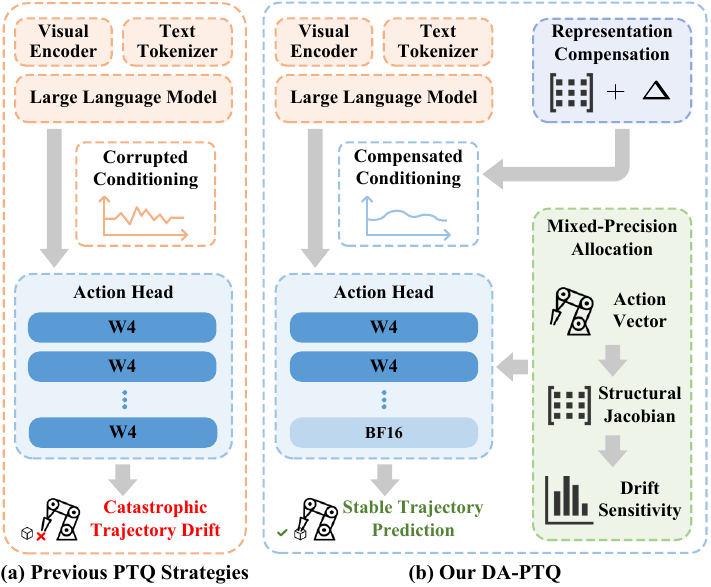}
    \caption{Comparison between previous PTQ strategies and our DA-PTQ. (a) Conventional uniform quantization distorts the conditioning interface, resulting in trajectory drift during sequential control. (b) DA-PTQ stabilizes execution through cross-space representation compensation and motion-driven mixed-precision allocation, thereby preserving kinematically sensitive components.}
    \label{intro}
\end{figure}

\section{Introduction}
Vision-Language-Action models (VLAs)~\cite{zhong2025survey} have emerged as a promising paradigm for embodied artificial intelligence, enabling robots to perform complex tasks based on visual observations and natural language instructions. By integrating large-scale vision encoders, language backbones, and action generation modules within a unified architecture, recent models such as RT-2~\cite{zitkovich2023rt} and OpenVLA~\cite{kim2024openvla} demonstrate strong generalization across a wide range of manipulation tasks. However, these capabilities come at the cost of substantial memory and computational overhead, often involving billions of parameters. Such requirements fundamentally conflict with the constraints of onboard robotic systems, where low latency, limited memory, and power efficiency are critical. Consequently, enabling efficient deployment of VLAs remains a central challenge for real-world embodied AI.

Model compression offers a natural pathway to address this challenge. Among various techniques, quantization is particularly attractive due to its ability to simultaneously reduce memory footprint and inference cost. While Quantization-Aware Training (QAT) \cite{jacob2018quantization} can recover accuracy through retraining, it is often impractical for large-scale VLA models due to the substantial computational cost and the limited availability of high-quality multimodal robotic data. In contrast, Post-Training Quantization (PTQ)~\cite{liu2021post} provides a lightweight alternative by calibrating models without retraining. Recent PTQ methods have achieved remarkable success in large language models and vision-language models. However, directly applying these techniques to VLAs often leads to severe performance degradation, and in some cases unstable control behaviors during sequential execution~\cite{zhang2026quantvla, xu2026qvla}.

We attribute this limitation to a fundamental mismatch between conventional quantization objectives and the sequential, control-sensitive nature of embodied decision-making. Existing PTQ methods typically assume that quantization errors are independent and locally bounded, such that minimizing layer-wise reconstruction error is sufficient to preserve downstream functionality (e.g., AWQ~\cite{lin2024awq}, GPTQ~\cite{frantar2022gptq}). While this assumption is reasonable for static generation tasks, it becomes inadequate in embodied control settings, where decisions are executed sequentially and coupled through system dynamics. In VLA models, quantization perturbs latent representations at the vision-language-to-action interface, and these perturbations become temporally coupled and progressively amplified over time~\cite{chen2024stepbaq, liu2026ttf}. As accumulated errors interact with robot dynamics and feedback control loops, they manifest as kinematic drift, i.e., the deviation between the executed trajectory and the nominal trajectory induced by quantization, ultimately resulting in significant performance degradation~\cite{park2025acg}.

Recent works have begun to explore quantization strategies tailored for VLAs. For example, SQAP-VLA~\cite{fang2025sqap} jointly optimizes quantization and token pruning, QuantVLA~\cite{zhang2026quantvla} stabilizes the perception-to-action interface via scale-calibrated adjustments, and QVLA~\cite{xu2026qvla} allocates bit-widths based on channel-wise action sensitivity. While these methods mitigate performance degradation to some extent, substantial gaps persist under aggressive compression. A key limitation is that existing approaches primarily rely on static or single-step approximations, failing to capture long-horizon error propagation and its interaction with robot dynamics. Consequently, they cannot effectively model or control trajectory-level error accumulation, leaving kinematic drift as a fundamental bottleneck in achieving an optimal efficiency-accuracy trade-off.

To address the above challenges, we propose Drift-Aware Post-Training Quantization (DA-PTQ), a training-free framework that explicitly models and mitigates kinematic drift, as conceptually illustrated in Figure~\ref{intro}. Instead of treating quantization as a static reconstruction problem, DA-PTQ formulates it as a drift-aware optimization process that explicitly accounts for both temporal accumulation and physical amplification of errors. Specifically, DA-PTQ consists of two complementary components. Cross-Space Representation Compensation mitigates structured distortions between multimodal representations and the action space via lightweight affine and low-rank transformations, improving action consistency under quantization. Motion-Driven Mixed-Precision Allocation further reduces long-horizon drift by assigning bit-widths based on trajectory-level motion error under resource constraints. Together, these components enable stable and efficient low-precision deployment without introducing additional inference overhead.

Our main contributions are summarized as follows:
\begin{itemize}
\item We present a systematic analysis of error accumulation in VLA quantization, identifying kinematic drift as a key bottleneck arising from the interplay between quantization perturbations and sequential embodied control.
\item We propose DA-PTQ, a training-free framework that reformulates quantization as a drift-aware optimization problem, integrating cross-space representation compensation and motion-driven precision allocation without incurring additional inference overhead.
\item Experimental results show that DA-PTQ reduces kinematic drift and achieves performance comparable to full-precision models under low-bit settings, enabling deployment on resource-constrained robotic platforms.
\end{itemize}

\section{Related Work}
\subsection{Vision-Language-Action Models}
Vision-Language-Action models (VLAs) have emerged as a dominant paradigm for generalist robotic control, learning direct mappings from visual observations and natural language instructions to executable motor commands. Early works such as RT-2~\cite{zitkovich2023rt} demonstrate that large-scale vision-language pretraining can be effectively transferred to robotic manipulation by treating actions as language tokens, thereby enabling strong cross-task generalization. OpenVLA~\cite{kim2024openvla} further advances this direction by open-sourcing a 7B-parameter model trained on diverse manipulation data, establishing a widely adopted benchmark for embodied control. More recently, Gemini Robotics~\cite{team2025gemini} extends this paradigm toward foundation-scale embodied agents by integrating large multimodal models with robotic control, demonstrating strong generalization and reasoning capabilities in real-world scenarios.

A parallel line of work models actions in the continuous domain using expressive generative decoders. Octo~\cite{team2024octo} and RDT-1B~\cite{liu2024rdt} adopt diffusion-based action heads to capture multimodal action distributions, while $\pi_0$~\cite{black2024pi_0} employs flow matching for high-frequency dexterous control. CogACT~\cite{li2024cogact} further integrates a Diffusion Transformer (DiT)~\cite{peebles2023scalable} action module into the VLA framework, showing that decoupling perception and action generation leads to more precise and temporally consistent control. More recently, OpenVLA-OFT~\cite{kim2025fine} introduces fine-tuning strategies that improve both efficiency and task success rates. Building on these advances, approaches such as DeepThinkVLA~\cite{yin2025deepthinkvla} incorporate explicit reasoning processes (e.g., chain-of-thought) into VLAs, enabling improved long-horizon planning and decision-making.

Despite these advances, a common trend across VLAs is the rapid scaling of both vision-language backbones and action generation modules. This scaling substantially increases memory footprint and inference latency, particularly for diffusion-based policies with iterative decoding. Consequently, deploying VLAs on resource-constrained robotic platforms remains a fundamental challenge, motivating the need for efficient model compression techniques such as quantization.

\subsection{Post-Training Quantization}

Post-training quantization (PTQ)~\cite{liu2021post} has emerged as a practical and widely adopted solution for compressing large-scale models without retraining, making it particularly suitable for deployment on resource-constrained hardware. Existing PTQ methods can be broadly categorized into reconstruction-based and transformation-based approaches, both aiming to preserve model performance under low-bit precision by maintaining feature-space fidelity.

Reconstruction-based methods, such as GPTQ~\cite{frantar2022gptq} and BRECQ \cite{li2021brecq}, minimize layer-wise or block-wise quantization error using second-order approximations, directly optimizing numerical reconstruction. In contrast, transformation-based methods reshape feature distributions to facilitate quantization. SmoothQuant~\cite{xiao2023smoothquant} mitigates activation outliers by shifting quantization difficulty to weights via channel-wise scaling, while AWQ~\cite{lin2024awq} preserves salient weight channels based on activation statistics. OmniQuant~\cite{shao2023omniquant} further extends this paradigm by jointly optimizing clipping thresholds and equivalent transformations to smooth both weight and activation distributions. Beyond unimodal settings, Q-VLM~\cite{wang2024q} adapts PTQ to multimodal architectures by mitigating cross-modal distortions. Despite their methodological differences, these approaches share a common underlying principle: improving quantization performance by preserving feature-space fidelity through numerical reconstruction or statistical alignment. This perspective is closely related to prior findings such as AdaIN~\cite{huang2017arbitrary}, which shows that aligning first- and second-order statistics via affine transformations can effectively correct distributional shifts, and LoRA~\cite{hu2022lora}, which suggests that effective adaptations often lie in low-dimensional subspaces, motivating compact low-rank corrections for structured feature distortions.

However, these methods fundamentally treat quantization as a static feature reconstruction problem, assuming that preserving local numerical fidelity is sufficient to maintain downstream performance. While this assumption holds in language and vision-language models, it breaks down in embodied control. In VLA models, quantization errors perturb latent representations at the perception-action boundary and are progressively amplified during sequential execution. As these errors accumulate and interact with robot dynamics, they manifest as kinematic drift, which cannot be captured by conventional PTQ objectives and ultimately limits the achievable efficiency-accuracy trade-off.

\subsection{Quantization for VLMs}
Recent works have begun to explore quantization strategies specifically tailored for VLAs. QuantVLA~\cite{zhang2026quantvla} identifies the perception-to-action interface as a critical bottleneck under quantization and proposes scale-calibrated affine adjustments to stabilize latent representations passed to the action generation module. QVLA~\cite{xu2026qvla} conducts a systematic sensitivity analysis of VLA architectures, revealing pronounced channel heterogeneity and introducing a channel-wise bit allocation strategy guided by action-space sensitivity scores derived from Taylor expansion. SQAP-VLA~\cite{fang2025sqap} further jointly optimizes quantization and token pruning within a unified framework, targeting redundancy in both the vision encoder and the language backbone.

The issue of error accumulation in sequential decision-making provides important context for understanding the limitations of these approaches. In imitation learning, Ross et al.~\cite{ross2011reduction} formally characterize compounding errors, showing that small perturbations can lead to quadratically increasing deviations over the task horizon. In robotics, manipulator Jacobian analysis~\cite{spong2012robot, siciliano2009robotics} shows that errors in proximal degrees of freedom are geometrically amplified along the kinematic chain, with Jacobian column norms quantifying the amplification factor of each dimension. This geometric asymmetry is intrinsic to serial manipulator structures and is independent of specific link parameters.

Despite these advances, existing VLA quantization methods exhibit two key limitations. First, channel sensitivity is generally estimated through static reconstruction metrics or single-step action deviations, which fail to capture how quantization errors are geometrically amplified and temporally accumulated during sequential execution.  Second, representational distortions at the vision-language-to-action interface are typically corrected at a coarse granularity, without explicitly modeling structured per-channel distributional shifts between full-precision and quantized feature spaces. DA-PTQ addresses these limitations through two complementary components. Cross-space representation compensation performs fine-grained correction of per-channel and cross-channel distributional distortions across modalities. Motion-driven mixed-precision allocation further incorporates kinematic error propagation to identify and preserve dimensions whose quantization errors most strongly accumulate into long-horizon trajectory drift.

\begin{figure*}[t]
    \centering
    \includegraphics[width=1\linewidth]{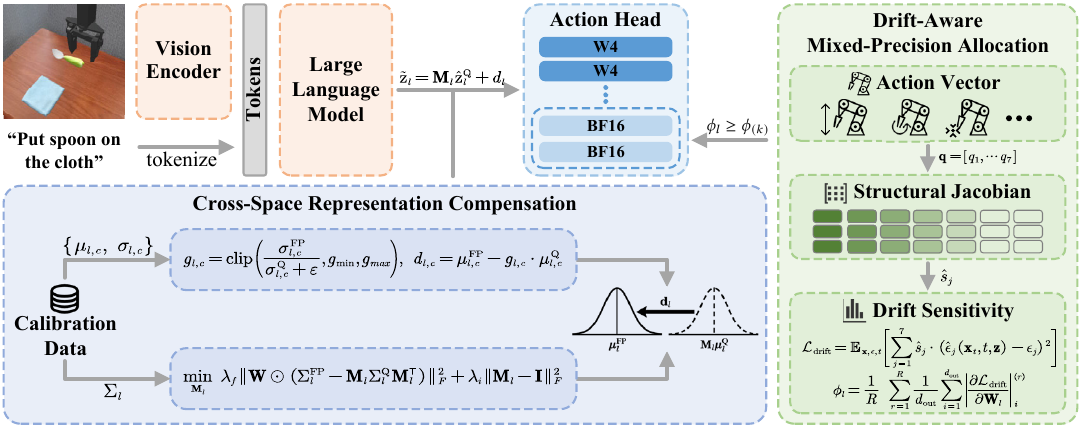}
    \caption{Overview of our proposed DA-PTQ framework. DA-PTQ enables efficient and robust diffusion-based VLA models through two key components: (1) Cross-Space Representation Compensation, which aligns distorted conditioning representations with their full-precision counterparts; and (2) Drift-Aware Mixed-Precision Allocation, which leverages structural Jacobian-based sensitivity to mitigate compounding trajectory drift.}
    \label{framework}
\end{figure*}

\section{DA-PTQ}
\subsection{Problem Formulation}
Vision-Language-Action models (VLAs) define a policy $\Pi_\theta$ that maps visual observations $\mathbf{V}_t$ and a language instruction $p$ to continuous actions $\mathbf{a}_t$:
\begin{equation}
    \Pi_\theta(\mathbf{a}_t \mid \mathbf{V}_t, p) = 
    \Psi_\theta\left(f_\theta(\mathbf{V}_t, p)\right),
\end{equation}
where $f_\theta$ denotes the vision-language backbone and $\Psi_\theta$ is the action decoder. The backbone encodes multimodal inputs into a latent representation $\mathbf{z}_t = f_\theta(\mathbf{V}_t, p)$, which serves as the conditioning signal for action generation.

In modern VLA architectures, $\Psi_\theta$ is often implemented as a diffusion-based policy, where actions are generated via iterative denoising conditioned on $\mathbf{z}_t$. While effective, this design poses unique challenges under Post-Training Quantization (PTQ), as latent-space quantization errors are repeatedly injected during decoding and further accumulate over time in sequential control. These challenges stem from two coupled mechanisms:

\vspace{1.5mm}
\noindent\textbf{Quantization Sensitivity at the Conditioning Interface.}
The perception-to-action interface $\mathbf{z}_t$ acts as a tightly coupled information bottleneck through which all task-relevant signals are conveyed to the action decoder. Under quantization, perturbations in $\mathbf{z}_t$ directly distort the conditioning distribution. Since the decoder depends solely on $\mathbf{z}_t$, these distortions propagate through all denoising steps without downstream correction, and are further amplified by the iterative diffusion process, resulting in structured distributional shifts even within a single action prediction.

\vspace{1.5mm}
\noindent\textbf{Temporal Error Accumulation in Sequential Control.}
Beyond single-step sensitivity, VLA policies operate in a closed-loop setting where actions affect future observations. Let $\boldsymbol{\epsilon}_t \in \mathbb{R}^7$ denote the action error at timestep $t$, primarily induced by quantization. The resulting end-effector deviation is:
\begin{equation}
    \delta \mathbf{e}_t = \mathbf{J}^{(t)} \boldsymbol{\epsilon}_t,
\end{equation}
where $\mathbf{J}^{(t)}$ is the manipulator Jacobian. Over a horizon $T$, the accumulated deviation becomes:
\begin{equation}
    \mathbf{E}_T = \sum_{t=1}^{T} \mathbf{J}^{(t)} \boldsymbol{\epsilon}_t.
\end{equation}
This shows that quantization errors are both temporally accumulated and geometrically amplified via $\|\mathbf{J}^{(t)}_{:,j}\|$, causing small per-step errors to induce significant long-horizon drift.

Taken together, quantization in VLAs induces both representation-level distortion and trajectory-level drift, which are tightly coupled yet not explicitly addressed by existing PTQ methods.

\subsection{Overview of DA-PTQ}
To address the coupled challenges of representation distortion and trajectory-level drift, we propose DA-PTQ, a streamlined calibration pipeline with no retraining overhead. As shown in Figure~\ref{framework}, it unifies cross-space representation compensation and drift-aware mixed-precision allocation into a three-stage process given a small calibration dataset.

First, we perform full-precision forward passes to collect calibration statistics. During this stage, we extract action trajectories to estimate the manipulator Jacobian and derive temporal drift propagation scores that characterize how errors accumulate over time. In parallel, we record activation statistics at the critical vision-language-to-action interfaces, capturing the reference distribution of conditioning representations.

Second, we apply cross-space representation compensation to correct distortion at the conditioning interface. After quantization, we perform a forward pass to measure activation shifts and solve for lightweight affine transformations that align quantized representations with their full-precision counterparts. These transformations are analytically merged into the quantized weights and biases, incurring no additional inference cost.

Finally, we conduct drift-aware mixed-precision allocation. Based on the compensated model, we perform a lightweight sensitivity analysis to evaluate the contribution of each layer to long-horizon error accumulation. This results in a layer-wise bit-width assignment that preserves higher precision for drift-sensitive components while aggressively compressing less critical layers.

The resulting DA-PTQ model achieves substantial compression while maintaining stable and accurate continuous control, effectively mitigating both conditioning distortion and long-horizon drift without incurring runtime overhead.

\subsection{Cross-Space Representation Compensation}
Quantization induces structured distributional distortions at the vision-language-to-action interface, where activation shifts corrupt the conditioning signal for action generation. To mitigate this, we introduce Cross-Space Representation Compensation (CSRC), which aligns quantized activations with their full-precision counterparts through a hierarchical compensation scheme.

Let $\mathbf{z}_{l,c}^{\text{FP}}$ and $\hat{\mathbf{z}}_{l,c}^{\text{Q}}$ denote the full-precision and quantized activations of channel $c$ at interface layer $l$. We first match their first- and second-order statistics computed on the calibration set. Specifically, we derive a scale factor to align the standard deviation:
\begin{equation}
    g_{l,c} = \mathrm{clip}\!\left(
    \frac{\sigma_{l,c}^{\text{FP}}}{\sigma_{l,c}^{\text{Q}} + \varepsilon},\ 
    g_{\min}, g_{\max}\right),
\end{equation}
and a bias term to restore the mean:
\begin{equation}
    d_{l,c} = \mu_{l,c}^{\text{FP}} - g_{l,c} \cdot \mu_{l,c}^{\text{Q}}.
\end{equation}
The corrected activation is:
\begin{equation}
    \tilde{z}_{l,c} = g_{l,c} \hat{z}_{l,c}^{\text{Q}} + d_{l,c}.
\end{equation}

While per-channel scaling corrects diagonal shifts, it fails to capture structured cross-channel distortion. To address this, we introduce a dense affine transformation $\mathbf{M}_l$ that aligns second-order statistics between full-precision and quantized activations. Let $\boldsymbol{\Sigma}_l^{\text{FP}}$ and $\boldsymbol{\Sigma}_l^{\text{Q}}$ denote their empirical covariance matrices. We solve:
\begin{equation}
    \min_{\mathbf{M}_l}\ 
    \lambda_f \left\| \mathbf{W} \odot 
    \left(\boldsymbol{\Sigma}_l^{\text{FP}} - 
    \mathbf{M}_l \boldsymbol{\Sigma}_l^{\text{Q}} 
    \mathbf{M}_l^\top\right) \right\|_F^2 
    + \lambda_i \left\| \mathbf{M}_l - \mathbf{I} \right\|_F^2,
\end{equation}
where $\mathbf{W}$ is a diagonal weight matrix derived from per-channel variance ratios, and $\lambda_i$ regularizes the solution toward identity to preserve stability.

To ensure efficiency, we parameterize $\mathbf{M}_l$ as a low-rank update to the identity:
\begin{equation}
    \mathbf{M}_l = \mathbf{I} + \mathbf{U}_l \mathbf{V}_l^\top, 
    \quad r \ll d,
\end{equation}
obtained via truncated SVD of the dense solution.

The fully corrected activation is given by:
\begin{equation}
    \tilde{\mathbf{z}}_l = \mathbf{M}_l \hat{\mathbf{z}}_l^{\text{Q}} 
    + \mathbf{d}_l,
\end{equation}
where the bias restores the mean of the aligned distribution. All compensation parameters are analytically folded into the quantized weights during calibration, incurring zero inference overhead.

\subsection{Drift-Aware Mixed-Precision Allocation}
To mitigate temporally accumulated errors in continuous control, we propose a Drift-Aware Mixed-Precision Allocation (DA-MPA) strategy that explicitly accounts for how quantization noise propagates and amplifies over long horizons.

In embodied control tasks, quantization errors introduced at each timestep accumulate through closed-loop interactions, resulting in trajectory drift and covariate shift. Directly modeling this process by unrolling environment dynamics is both computationally prohibitive and non-differentiable. To address this, we adopt an analytical surrogate that approximates temporal drift via single-step spatial error propagation. The action vector is defined as a 7-dimensional Cartesian increment:
\begin{equation}
    \mathbf{a}_t = [\Delta x, \Delta y, \Delta z, \Delta r_x, \Delta r_y, \Delta r_z, \Delta g] \in \mathbb{R}^7.
\end{equation}

Under the small-angle assumption ($\|\Delta \mathbf{r}\| \ll 1$), rotational components can be approximated as angular deviations:
\begin{equation}
    \Delta r_i \approx \delta \theta_i.
\end{equation}

For typical tabletop manipulation, drift is dominated by motion in the horizontal plane. We therefore project the action into three principal components corresponding to planar translation and rotation.

To model error accumulation, we reinterpret the action dimensions as joint increments of a virtual planar serial chain, $\mathbf{q} = [q_1, \dots, q_7]$. The absolute orientation of the $j$-th segment accumulates upstream perturbations:
\begin{equation}
    \theta_j = \sum_{i=1}^{j} q_i,
\end{equation}
which provides a differentiable proxy for how per-dimension errors propagate and accumulate, mimicking long-horizon drift behavior.

We quantify how perturbations in each action dimension affect the end-effector by deriving the structural Jacobian $\mathbf{J} \in \mathbb{R}^{3 \times 7}$ of the virtual chain:
\begin{equation}
J_x^{(j)} = -\sum_{k=j}^{6} \sin \theta_k, \ \ \
J_y^{(j)} = \sum_{k=j}^{6} \cos \theta_k, \ \ \
J_\theta^{(j)} = 1.
\end{equation}

Stacking these components yields:
\begin{equation}
    \mathbf{J} =
    \begin{bmatrix}
        J_x^{(1)} & \cdots & J_x^{(7)} \\
        J_y^{(1)} & \cdots & J_y^{(7)} \\
        1 & \cdots & 1
    \end{bmatrix}.
\end{equation}

This structure naturally captures error amplification: earlier dimensions influence more downstream segments, resulting in larger column norms, i.e., $\|\mathbf{J}_{:,j}\| > \|\mathbf{J}_{:,j+1}\|$. Thus, the Jacobian encodes the intrinsic topology of drift propagation without requiring task-specific supervision.

To isolate the contribution of each dimension to overall drift, we compute the damped least-squares pseudo-inverse:
\begin{equation}
    \mathbf{J}^+ = \mathbf{J}^\top \left( \mathbf{J} \mathbf{J}^\top + \lambda \mathbf{I}_3 \right)^{-1},
\end{equation}
where $\lambda > 0$ ensures numerical stability.

We further introduce axis-dependent weights to balance translational and rotational sensitivities. The drift propagation score for dimension $j$ is defined as:
\begin{equation}
    s_j = \mathbb{E}_{\mathbf{a} \sim \mathcal{D}} \left[ \sum_{c \in \{x, y, \theta\}} w_c \left| J^+_{j,c} \right| \right].
\end{equation}

We normalize these scores to obtain drift sensitivity weights:
\begin{equation}
    \hat{s}_j = \frac{s_j}{\frac{1}{7} \sum_{i=1}^{7} s_i}.
\end{equation}

These weights quantify how strongly errors in each action dimension contribute to long-horizon drift.

We incorporate the drift sensitivity weights into the calibration objective to penalize quantization noise that disproportionately amplifies drift:
\begin{equation}
\mathcal{L}_{\text{drift}} =
\mathbb{E}_{\mathbf{x}, \boldsymbol{\epsilon}, t}
\left[
\sum_{j=1}^{7}
\hat{s}_j \cdot
\left(
\hat{\epsilon}_j(\mathbf{x}_t, t, \mathbf{z}) - \epsilon_j
\right)^2
\right].
\end{equation}

Under this objective, gradients are automatically reweighted to emphasize dimensions with high drift sensitivity. For each quantizable layer $l$, we compute its drift sensitivity score by averaging gradient magnitudes over $R$ calibration steps:
\begin{equation}
    \phi_l =
    \frac{1}{R}
    \sum_{r=1}^{R}
    \frac{1}{d_{\text{out}}}
    \sum_{i=1}^{d_{\text{out}}}
    \left|
    \frac{\partial \mathcal{L}_{\text{drift}}}{\partial \mathbf{W}_l}
    \right|_i^{(r)}.
\end{equation}

Layers are then ranked according to $\phi_l$. To tightly control temporal drift, the top $k\%$ most sensitive layers are retained in high precision (BF16), while the remaining layers are quantized to low bit-width:
\begin{equation}
    b_l =
    \begin{cases}
    \text{BF16}, & \text{if } \phi_l \geq \phi_{(k)}, \\
    \text{W4}, & \text{otherwise}.
    \end{cases}
\end{equation}

\subsection{Summary of the DA-PTQ Pipeline}
We summarize DA-PTQ as a three-stage calibration procedure in Algorithm~\ref{alg:da_ptq}. The pipeline is fully training-free and requires only forward passes and lightweight gradient accumulation. Given a pretrained VLA model and a small calibration dataset, the procedure first performs drift profiling by collecting full-precision activation statistics and estimating both per-dimension drift sensitivity and layer-wise impact on error accumulation.  Next, a drift-aware mixed-precision configuration is determined by retaining the most sensitive layers in high precision while aggressively quantizing the rest. Finally, cross-space representation compensation is calibrated on the quantized model by aligning activation distributions and folding the resulting affine transformations into the weights, incurring zero inference overhead.

\begin{algorithm}[t]
\caption{Drift-Aware Post-Training Quantization (DA-PTQ)}
\label{alg:da_ptq}
\textbf{Input:} Pretrained VLA model with DiT action head $\Psi_\theta$, calibration dataset $\mathcal{D}$, retained BF16 ratio $k\%$ \\
\textbf{Output:} Quantized VLA model with W4/BF16 mixed precision and folded compensation
\begin{algorithmic}[1]
    \Statex \textcolor{gray}{\# Stage 1: Drift Profiling}
    \For{each batch in $\mathcal{D}$}
        \State Accumulate full-precision statistics $\boldsymbol{\mu}_l^{\text{FP}}$, $\boldsymbol{\Sigma}_l^{\text{FP}}$ at the perception-action interface
        \State Compute structural Jacobian $\mathbf{J}$ and per-dimension drift sensitivities $\hat{s}_j$
        \State Estimate layer-wise drift sensitivities $\phi_l$ under $\mathcal{L}_{\text{drift}}$
    \EndFor

    \Statex \textcolor{gray}{\# Stage 2: Cross-Space Representation Compensation}
    \State Quantize the model with the initial calibration configuration
    \For{each batch in $\mathcal{D}$}
        \State Accumulate quantized statistics $\boldsymbol{\mu}_l^{\text{Q}}$, $\boldsymbol{\Sigma}_l^{\text{Q}}$
    \EndFor
    \For{each interface layer $l$}
        \State Solve for affine matrix $\mathbf{M}_l$ and bias $\mathbf{d}_l$ to align quantized statistics with full-precision statistics
        \State Fold $\mathbf{M}_l$ and $\mathbf{d}_l$ into the quantized weights of layer $l$
    \EndFor

    \Statex \textcolor{gray}{\# Stage 3: Drift-Aware Mixed-Precision Allocation}
    \For{each layer $l \in \Psi_\theta$}
        \State $b_l \leftarrow \text{BF16}$ if $\phi_l \ge \phi_{(k)}$ else $\text{W4}$
    \EndFor
    \State Apply the bit-width map $\{b_l\}$ to obtain the final quantized model

    \State \Return Compressed VLA model
\end{algorithmic}
\end{algorithm}

\section{Experiment}

\subsection{Experiment Setup}
\noindent\textbf{Benchmark and Evaluation.}
We evaluate DA-PTQ on SimplerEnv, a standardized simulation benchmark for robotic manipulation. To ensure comprehensive evaluation, we consider two distinct embodiments: the WidowX robot and the Google Robot. On WidowX, we report performance across four standard manipulation tasks. On the Google Robot, we focus on zero-shot cross-domain generalization, evaluating under both Visual Matching and the more challenging Variant Aggregation settings.

\vspace{1.5mm}
\noindent\textbf{Calibration Process and Dataset.}
As a post-training quantization framework, DA-PTQ requires only a small calibration dataset to estimate drift sensitivities and compute cross-space compensation parameters, without any model fine-tuning. We construct the calibration set using 512 representative trajectories sampled from the training split of BridgeData V2~\cite{walke2023bridgedata}. Notably, this single calibration set is used to quantize the model for both in-domain WidowX evaluation and zero-shot cross-domain Google Robot evaluation. This design provides a stringent test of whether the learned drift-aware and distribution-aligned quantization parameters generalize beyond the calibration domain.

\vspace{1.5mm}
\noindent\textbf{Backbone and Baseline.}
We apply DA-PTQ with W4A8 quantization to CogACT, a state-of-the-art diffusion-based VLA model. We compare against the following strong baselines:
\begin{itemize}
    \item \textbf{CogACT (FP)}: The full-precision model without quantization.
    \item \textbf{VLA-Cache}: An inference acceleration method that caches intermediate representations but does not perform weight or activation quantization.
    \item \textbf{QuantVLA}: A recent VLA-specific quantization method based on conventional sensitivity metrics without explicit modeling of temporal drift.
\end{itemize}
All metrics, including success rate (\%), memory reduction (\%), and inference speedup (\%), are measured under identical hardware settings to ensure fair comparison.

\subsection{Comparison with Baselines}
We compare DA-PTQ with strong baselines to systematically evaluate both computational efficiency and task performance under low-bit quantization. Our goal is to examine not only the standalone gains in efficiency, but also how well each method preserves control fidelity under aggressive compression.

\begin{table}[b]
\centering
\caption{Efficiency-performance trade-off on SimplerEnv.}
\vspace{-1mm}
\label{tab:tradeoffs}
\renewcommand{\arraystretch}{1.1}
\resizebox{\linewidth}{!}{
\begin{tabular}{lcccc}
\toprule
Method & Sources & \makecell{Success \\ Rate ($\uparrow$)} & \makecell{Memory \\ Reduction ($\uparrow$)} & \makecell{Speedup ($\uparrow$)} \\
\midrule
VLA-Cache & NeurIPS'25 & 46.8 & 0.0 & 36.7 \\
QuantVLA & CVPR'26 & 43.5 & 42.7 & 55.8 \\
\textbf{Ours} & \textbf{---} & \textbf{48.9} & \textbf{42.5} & \textbf{54.8} \\
\bottomrule
\end{tabular}}
\end{table}

\vspace{1.5mm}
\noindent\textbf{Efficiency-Performance Trade-off.}
Table~\ref{tab:tradeoffs} summarizes the trade-off between efficiency and task success. DA-PTQ achieves a favorable balance, delivering a 42.5\% memory reduction and a 54.8\% inference speedup while maintaining strong task performance. Compared to existing approaches, our method achieves nearly the same level of efficiency as aggressive low-bit quantization, yet avoids the significant degradation in control quality. Although QuantVLA attains slightly higher efficiency, with 42.7\% memory reduction and 55.8\% speedup, it suffers a substantial 5.4\% drop in success rate, indicating a clear loss in control fidelity. This gap highlights that optimizing for efficiency alone can lead to severe degradation in long-horizon decision-making. In contrast, VLA-Cache provides moderate acceleration at 36.7\% speedup but fails to reduce memory consumption, which fundamentally limits its deployment in memory-constrained scenarios. These results demonstrate that naive quantization or caching strategies are insufficient to balance efficiency and performance in embodied control.

\begin{table*}[t] 
\centering 
\caption{In-domain performance on WidowX under the SimplerEnv Visual Matching setting.}
\label{tab:accuracy}
\renewcommand{\arraystretch}{1.02}
\setlength{\tabcolsep}{10pt}
\resizebox{\linewidth}{!}{
\begin{tabular}{llc|cccc|c}
\toprule
\multirow{4}{*}{WidowX Robot} & \multirow{4}{*}{Methods} & \multirow{4}{*}{Sources} & \multicolumn{4}{c|}{Tasks} & \multirow{4}{*}{\makecell{Average \\ Success \\ Rate ($\uparrow$)}} \\
\cmidrule(lr){4-7}
& & & \makecell{Put Spoon \\ on Towel} & \makecell{Put Carrot \\ on Plate} & \makecell{Stack Green \\ Block on \\ Yellow Block} & \makecell{Put Eggplant \\ in Yellow \\ Basket}  \\
\midrule
\multirow{4}{*}{\makecell[l]{SIMPLER \\ (Visual Matching)}}
& CogACT (FP) & Arxiv'24 & 71.7 & 50.8 & 15.0 & 67.5 & 51.3 \\
& VLA-Cache & NeurIPS'25 & 78.3 & 39.1 & 17.4 & 52.2 & 46.8 \\
& QuantVLA & CVPR'26 & 47.8 & 39.1 & 17.4 & 69.6 & 43.5 \\
& \textbf{Ours} & --- & \textbf{65.2} & \textbf{52.2} & \textbf{17.4} & \textbf{60.9} & \textbf{48.9} \\
\bottomrule
\end{tabular}}
\end{table*}

\begin{table*}[t]
\centering
\caption{Cross-domain performance on Google Robot under SimplerEnv Visual Matching and Variant Aggregation settings.}
\label{tab:google_robot}
\renewcommand{\arraystretch}{1.02}
\setlength{\tabcolsep}{10pt} 
\resizebox{\linewidth}{!}{
\begin{tabular}{llc|cccc|c}
\toprule
\multirow{3}{*}{Google Robot} & \multirow{3}{*}{Methods} & \multirow{3}{*}{Sources} & \multicolumn{4}{c|}{Tasks} & \multirow{3}{*}{\makecell{Average \\ Success \\ Rate ($\uparrow$)}} \\
\cmidrule(lr){4-7}
& & & \makecell{Pick Coke \\ Can} & \makecell{Move \\  Near} & \makecell{Open/Close \\ Drawer} & \makecell{Open Top Drawer \\ and Place Apple} \\
\midrule
\multirow{4}{*}{\makecell[l]{SIMPLER \\ (Visual Matching)}}
 & CogACT (FP) & Arxiv'24 & 91.3 & 85.0 & 71.8 & 50.9  & 74.8 \\
 & VLA-Cache & NeurIPS'25 & 92.0 & 83.3 & 70.5 & 51.6 & 74.4 \\
 & QuantVLA & CVPR'26 & 87.6 & 81.7 & 55.1 & 38.0 & 65.6 \\
 & \textbf{Ours} & --- & \textbf{92.4} & \textbf{87.9} & \textbf{58.3} & \textbf{35.2} & \textbf{68.5} \\
\midrule
\multirow{4}{*}{\makecell[l]{SIMPLER \\ (Variant Aggregation)}}
 & CogACT (FP) & Arxiv'24 & 89.6 & 80.8 & 28.3 & 46.6  & 61.3 \\
 & VLA-Cache & NeurIPS'25  & 91.7 & 79.3 & 32.5 & 45.8 & 62.3 \\
 & QuantVLA & CVPR'26  & 84.9 & 76.7 & 20.3 & 15.8 & 49.4 \\
 & \textbf{Ours} & --- & \textbf{87.5} & \textbf{74.5} & \textbf{20.1} & \textbf{24.7} & \textbf{51.7} \\
\bottomrule
\end{tabular}}
\end{table*}

\vspace{1.5mm}
\noindent\textbf{In-Domain Performance on WidowX Robot.}
Table~\ref{tab:accuracy} reports the performance comparison on the WidowX robot under the SimplerEnv Visual Matching setting. DA-PTQ achieves an average success rate of 48.9\%, outperforming both QuantVLA at 43.5\% and VLA-Cache at 46.8\%, while substantially narrowing the gap to the full-precision CogACT. This indicates that DA-PTQ preserves the majority of the original model capability despite aggressive compression. A closer examination across individual tasks reveals consistent improvements on scenarios that require precise spatial coordination. For example, DA-PTQ achieves 65.2\% on \textit{Put Spoon on Towel} compared to 47.8\% for QuantVLA, and 52.2\% on \textit{Put Carrot on Plate} compared to 39.1\%. Meanwhile, it maintains competitive performance on tasks involving physical interaction, such as block stacking, where contact dynamics play a larger role. These results suggest that DA-PTQ effectively preserves fine-grained action control under quantization. This improvement can be attributed to the drift-aware allocation strategy, which explicitly prioritizes dimensions that contribute most to long-horizon error accumulation. By protecting these critical components, the model maintains stable execution trajectories even under reduced precision.

\vspace{1.5mm}
\noindent\textbf{Cross-Domain Generalization on Google Robot.}
We further evaluate robustness under cross-domain transfer to unseen robot embodiments. As shown in Table~\ref{tab:google_robot}, DA-PTQ achieves average success rates of 68.5\% in Visual Matching and 51.7\% in the more challenging Variant Aggregation setting, consistently outperforming QuantVLA across both regimes. The advantage becomes more evident under stronger distribution shift. QuantVLA exhibits significant degradation on tasks involving articulated objects or long-horizon planning, indicating sensitivity to accumulated errors. In contrast, DA-PTQ maintains stable performance on simpler tasks and reduces failures on more challenging scenarios, demonstrating improved robustness. Although a gap remains compared to the full-precision model, DA-PTQ avoids the severe collapse typically observed under aggressive low-bit quantization. This robustness is enabled by cross-space representation compensation, which aligns quantized activation statistics with their full-precision counterparts at the vision-action interface. As a result, the conditioning signal remains stable under domain shift, preventing error propagation into downstream action generation.

Across both in-domain and cross-domain settings, DA-PTQ consistently achieves the best balance between efficiency and performance. Compared to existing approaches, it preserves high task success while maintaining strong compression and acceleration. This advantage arises from the complementary design of the framework. Drift-aware mixed-precision allocation controls long-horizon error accumulation by protecting critical dimensions, while cross-space representation compensation corrects distributional distortions at the conditioning interface. Together, these components enable DA-PTQ to deliver robust embodied control under low-bit quantization without introducing additional runtime overhead.

\begin{figure}[b]
    \centering
    \includegraphics[width=\linewidth]{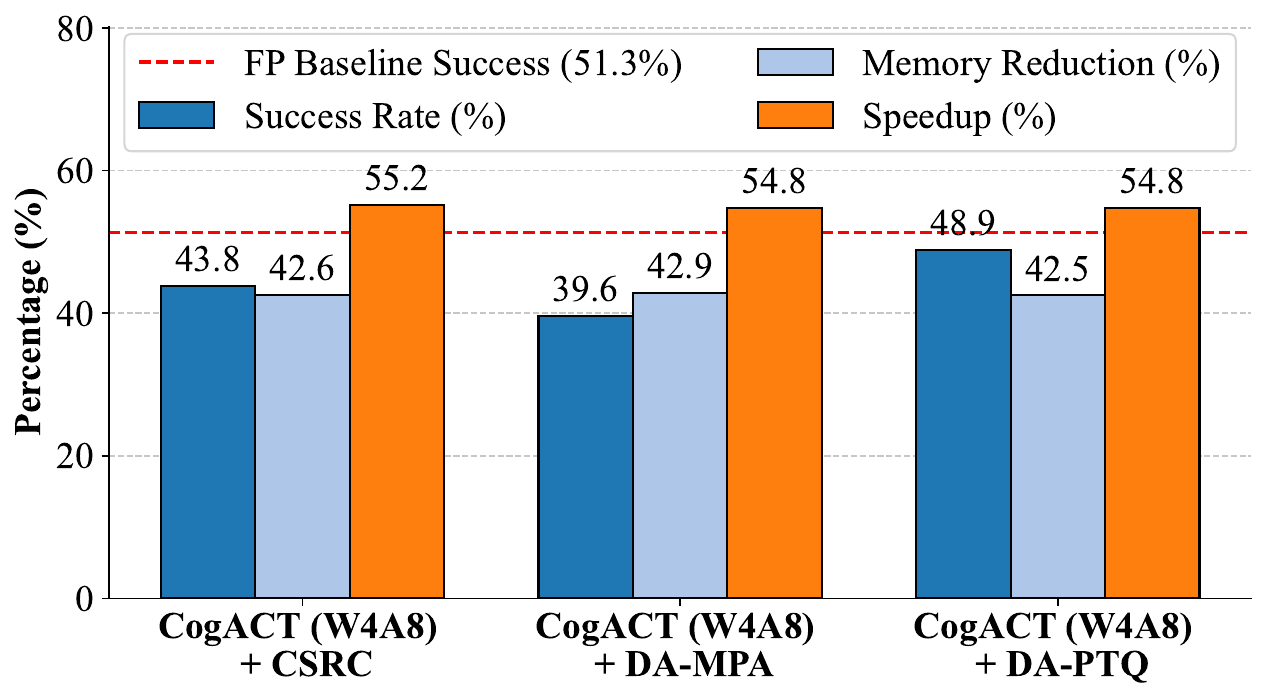} 
    \vspace{-6mm}
    \caption{Efficiency-performance trade-off across ablation variants on SimplerEnv.}
    \label{fig:ablation_tradeoffs}
\end{figure}

\begin{table*}[t] 
\centering 
\caption{Ablation study on WidowX under the SimplerEnv Visual Matching setting.}
\label{tab:ablation}
\renewcommand{\arraystretch}{1.02}
\setlength{\tabcolsep}{5pt}
\resizebox{\linewidth}{!}{
\begin{tabular}{ll|cccc|ccc}
\toprule
\multirow{4}{*}{WidowX Robot} & \multirow{4}{*}{Variants} & \multicolumn{4}{c|}{Tasks} & \multirow{4}{*}{\makecell{Average \\ Success \\ Rate ($\uparrow$)}} & \multirow{4}{*}{\makecell{Memory \\ Reduction ($\uparrow$)}} & \multirow{4}{*}{Speedup ($\uparrow$)} \\
\cmidrule(lr){3-6}
& &  \makecell{Put Spoon \\ on Towel} & \makecell{Put Carrot \\ on Plate} & \makecell{Stack Green \\ Block on \\ Yellow Block} & \makecell{Put Eggplant \\ in Yellow \\ Basket}  \\
\midrule
\multirow{4}{*}{\makecell[l]{SIMPLER \\ (Visual Matching)}}
& CogACT (FP) & 71.7 & 50.8 & 15.0 & 67.5 & --- & --- & --- \\
& \ \ + W4A8 + CSRC & 66.7 & 41.7 & 12.5 & 54.2 & 43.8 & 42.6 & 55.2 \\
& \ \ + W4A8 + DA-MPA & 50.0 & 45.8 & 12.5 & 50.0 & 39.6 & 42.9 & 54.8 \\
& \textbf{\ \ + W4A8 + DA-PTQ} & 65.2 & 52.2 & 17.4 & 60.9 & 48.9 & 42.5 & 54.8 \\
\bottomrule
\end{tabular}}
\end{table*}

\subsection{Ablation Study}
To systematically evaluate the contributions of each component, we conduct an ablation study on the WidowX robot under the Visual Matching setting. We isolate the effects of Cross-Space Representation Compensation (CSRC) and Drift-Aware Mixed-Precision Allocation (DA-MPA), and report both efficiency and performance in Table~\ref{tab:ablation}, with a visual comparison in Figure~\ref{fig:ablation_tradeoffs}. This study aims to clarify how each module contributes to preserving control performance under model quantization.

\vspace{1.5mm}
\noindent\textbf{Impact of Cross-Space Representation Compensation.}
Applying Cross-Space Representation Compensation alone to a uniformly quantized model with W4A8 precision yields an average success rate of 43.8\%. Although this remains below the full-precision baseline of 51.3\%, it already provides a clear improvement over naive quantization. The gain is particularly evident on tasks that require accurate perception-to-action alignment, such as \textit{Put Spoon on Towel} with 66.7\% success and \textit{Put Eggplant in Yellow Basket} with 54.2\%. These tasks depend heavily on precise conditioning signals, and the results indicate that correcting distributional distortion at the interface is essential for maintaining semantic consistency. By aligning quantized activation statistics with their full-precision counterparts, this component stabilizes the input to the diffusion decoder and reduces early-stage divergence during action generation.

\vspace{1.5mm}
\noindent\textbf{Impact of Drift-Aware Mixed-Precision Allocation.}
When applying drift-aware mixed-precision allocation alone, the model achieves an average success rate of 39.6\%. Although lower than the representation compensation variant, this result reflects a distinct and complementary effect. As illustrated in Figure~\ref{fig:ablation_tradeoffs}, this variant maintains strong efficiency while selectively improving performance on tasks sensitive to accumulated control errors. The underlying reason is that allocating higher precision to drift-sensitive layers effectively limits the propagation of quantization noise. Even when the conditioning representation is not fully corrected, the kinematic-aware allocation strategy constrains spatial error amplification along the execution trajectory, thereby improving stability in long-horizon control.

\vspace{1.5mm}
\noindent\textbf{Synergy of the Complete Framework.}
Combining both components yields the full DA-PTQ framework, which achieves the highest average success rate of 48.9\% while preserving the same level of efficiency, including 42.5\% memory reduction and 54.8\% inference speedup. As illustrated in Figure~\ref{fig:ablation_tradeoffs}, the complete model lies on the optimal Pareto frontier, outperforming both individual components without sacrificing efficiency. This result demonstrates a clear synergy between the two modules. Representation compensation corrects distributional distortion at the conditioning interface and preserves semantic alignment, while drift-aware allocation suppresses error accumulation during execution. By jointly addressing representation-level misalignment and trajectory-level drift, the combined framework provides a more complete and robust solution for embodied control under low-bit quantization.

\section{Conclusion}
We present DA-PTQ, a training-free post-training quantization framework for vision-language-action models. We identify temporal error accumulation as a key challenge in VLA quantization, where perturbations at the vision-language-to-action interface are progressively amplified during sequential control, leading to kinematic drift. To address this issue, DA-PTQ formulates quantization as a drift-aware optimization problem with two complementary components: cross-space representation compensation, which corrects structured distortions across modalities, and motion-driven mixed-precision allocation, which mitigates trajectory-level error accumulation. Both components are applied during calibration and introduce no additional inference overhead. Extensive experiments demonstrate that our DA-PTQ significantly reduces kinematic drift and achieves performance comparable to full-precision models under low-bit settings, enabling efficient deployment on resource-constrained robotic platforms.

Despite these advances, our approach relies on approximations of error propagation and simplified kinematic, which may not fully capture complex dynamics in highly nonlinear or contact-rich scenarios. Moreover, the calibration process assumes a fixed data distribution, potentially limiting robustness under significant domain shifts. Future work will extend drift-aware quantization to broader VLA architectures and develop more adaptive calibration strategies for diverse real-world environments.

\bibliographystyle{ACM-Reference-Format}
\bibliography{references}

@String{Computer = "{IEEE} Computer" }

@String{Chelsea = "Chelsea" }

@String{Springer = "Springer-Verlag" }

@article{black2024pi_0,
  title={$\pi_0 $: A Vision-Language-Action Flow Model for General Robot Control},
  author={Black, Kevin and Brown, Noah and Driess, Danny and Esmail, Adnan and Equi, Michael and Finn, Chelsea and Fusai, Niccolo and Groom, Lachy and Hausman, Karol and Ichter, Brian and others},
  journal={arXiv preprint arXiv:2410.24164},
  year={2024}
}

@article{chen2024stepbaq,
  title={Stepbaq: Stepping backward as correction for quantized diffusion models},
  author={Chen, Yi-Chung and Huang, Zhi-Kai and Chen, Jing-Ren},
  journal={Advances in Neural Information Processing Systems},
  pages={54054--54078},
  year={2024}
}

@article{fang2025sqap,
  title={Sqap-vla: A synergistic quantization-aware pruning framework for high-performance vision-language-action models},
  author={Fang, Hengyu and Liu, Yijiang and Du, Yuan and Du, Li and Yang, Huanrui},
  journal={arXiv preprint arXiv:2509.09090},
  year={2025}
}

@article{frantar2022gptq,
  title={Gptq: Accurate post-training quantization for generative pre-trained transformers},
  author={Frantar, Elias and Ashkboos, Saleh and Hoefler, Torsten and Alistarh, Dan},
  journal={arXiv preprint arXiv:2210.17323},
  year={2022}
}

@article{hu2022lora,
  title={Lora: Low-rank adaptation of large language models},
  author={Hu, Edward J and Shen, Yelong and Wallis, Phillip and Allen-Zhu, Zeyuan and Li, Yuanzhi and Wang, Shean and Wang, Liang and Chen, Weizhu and others},
  journal={Proceedings of the International Conference on Learning Representations},
  pages={1-12},
  year={2022}
}

@inproceedings{huang2017arbitrary,
  title={Arbitrary style transfer in real-time with adaptive instance normalization},
  author={Huang, Xun and Belongie, Serge},
  booktitle={Proceedings of the IEEE International Conference on Computer Vision},
  pages={1501--1510},
  year={2017}
}

@inproceedings{jacob2018quantization,
  title={Quantization and training of neural networks for efficient integer-arithmetic-only inference},
  author={Jacob, Benoit and Kligys, Skirmantas and Chen, Bo and Zhu, Menglong and Tang, Matthew and Howard, Andrew and Adam, Hartwig and Kalenichenko, Dmitry},
  booktitle={Proceedings of the IEEE Conference on Computer Vision and Pattern Recognition},
  pages={2704--2713},
  year={2018}
}

@article{kim2024openvla,
  title={Openvla: An open-source vision-language-action model},
  author={Kim, Moo Jin and Pertsch, Karl and Karamcheti, Siddharth and Xiao, Ted and Balakrishna, Ashwin and Nair, Suraj and Rafailov, Rafael and Foster, Ethan and Lam, Grace and Sanketi, Pannag and others},
  journal={arXiv preprint arXiv:2406.09246},
  year={2024}
}

@article{kim2025fine,
  title={Fine-tuning vision-language-action models: Optimizing speed and success},
  author={Kim, Moo Jin and Finn, Chelsea and Liang, Percy},
  journal={arXiv preprint arXiv:2502.19645},
  year={2025}
}

@article{li2021brecq,
  title={Brecq: Pushing the limit of post-training quantization by block reconstruction},
  author={Li, Yuhang and Gong, Ruihao and Tan, Xu and Yang, Yang and Hu, Peng and Zhang, Qi and Yu, Fengwei and Wang, Wei and Gu, Shi},
  journal={arXiv preprint arXiv:2102.05426},
  year={2021}
}

@article{li2024cogact,
  title={Cogact: A foundational vision-language-action model for synergizing cognition and action in robotic manipulation},
  author={Li, Qixiu and Liang, Yaobo and Wang, Zeyu and Luo, Lin and Chen, Xi and Liao, Mozheng and Wei, Fangyun and Deng, Yu and Xu, Sicheng and Zhang, Yizhong and others},
  journal={arXiv preprint arXiv:2411.19650},
  year={2024}
}

@article{lin2024awq,
  title={Awq: Activation-aware weight quantization for on-device llm compression and acceleration},
  author={Lin, Ji and Tang, Jiaming and Tang, Haotian and Yang, Shang and Chen, Wei-Ming and Wang, Wei-Chen and Xiao, Guangxuan and Dang, Xingyu and Gan, Chuang and Han, Song},
  journal={Proceedings of Machine Learning and Systems},
  volume={6},
  pages={87--100},
  year={2024}
}

@article{liu2021post,
  title={Post-training quantization for vision transformer},
  author={Liu, Zhenhua and Wang, Yunhe and Han, Kai and Zhang, Wei and Ma, Siwei and Gao, Wen},
  journal={Advances in Neural Information Processing Systems},
  pages={28092--28103},
  year={2021}
}

@article{liu2024rdt,
  title={Rdt-1b: a diffusion foundation model for bimanual manipulation},
  author={Liu, Songming and Wu, Lingxuan and Li, Bangguo and Tan, Hengkai and Chen, Huayu and Wang, Zhengyi and Xu, Ke and Su, Hang and Zhu, Jun},
  journal={arXiv preprint arXiv:2410.07864},
  year={2024}
}

@inproceedings{liu2026ttf,
  title={Ttf-vla: Temporal token fusion via pixel-attention integration for vision-language-action models},
  author={Liu, Chenghao and Zhang, Jiachen and Li, Chengxuan and Zhou, Zhimu and Wu, Shixin and Huang, Songfang and Duan, Huiling},
  booktitle={Proceedings of the AAAI Conference on Artificial Intelligence},
  pages={18452--18459},
  year={2026}
}

@article{park2025acg,
  title={ACG: Action Coherence Guidance for Flow-based VLA models},
  author={Park, Minho and Kim, Kinam and Hyung, Junha and Jang, Hyojin and Jin, Hoiyeong and Yun, Jooyeol and Lee, Hojoon and Choo, Jaegul},
  journal={arXiv preprint arXiv:2510.22201},
  year={2025}
}

@inproceedings{peebles2023scalable,
  title={Scalable diffusion models with transformers},
  author={Peebles, William and Xie, Saining},
  booktitle={Proceedings of the IEEE International Conference on Computer Vision},
  pages={4195--4205},
  year={2023}
}

@inproceedings{ross2011reduction,
  title={A reduction of imitation learning and structured prediction to no-regret online learning},
  author={Ross, St{\'e}phane and Gordon, Geoffrey and Bagnell, Drew},
  booktitle={Proceedings of the International Conference on Artificial Intelligence and Statistics},
  pages={627--635},
  year={2011}
}

@article{team2025gemini,
  title={Gemini robotics: Bringing ai into the physical world},
  author={Team, Gemini Robotics and Abeyruwan, Saminda and Ainslie, Joshua and Alayrac, Jean-Baptiste and Arenas, Montserrat Gonzalez and Armstrong, Travis and Balakrishna, Ashwin and Baruch, Robert and Bauza, Maria and Blokzijl, Michiel and others},
  journal={arXiv preprint arXiv:2503.20020},
  year={2025}
}

@article{team2024octo,
  title={Octo: An open-source generalist robot policy},
  author={Team, Octo Model and Ghosh, Dibya and Walke, Homer and Pertsch, Karl and Black, Kevin and Mees, Oier and Dasari, Sudeep and Hejna, Joey and Kreiman, Tobias and Xu, Charles and others},
  journal={arXiv preprint arXiv:2405.12213},
  year={2024}
}

@article{shao2023omniquant,
  title={Omniquant: Omnidirectionally calibrated quantization for large language models},
  author={Shao, Wenqi and Chen, Mengzhao and Zhang, Zhaoyang and Xu, Peng and Zhao, Lirui and Li, Zhiqian and Zhang, Kaipeng and Gao, Peng and Qiao, Yu and Luo, Ping},
  journal={arXiv preprint arXiv:2308.13137},
  year={2023}
}

@book{siciliano2009robotics,
  title={Robotics: modelling, planning and control},
  author={Siciliano, Bruno and Sciavicco, Lorenzo and Villani, Luigi and Oriolo, Giuseppe},
  year={2009},
  publisher={Springer}
}

@book{spong2012robot,
  title={Robot modeling and control},
  author={Spong, Mark W and Hutchinson, Seth and Vidyasagar, Mathukumalli},
  year={2020},
  publisher={Wiley}
}

@inproceedings{walke2023bridgedata,
  title={Bridgedata v2: A dataset for robot learning at scale},
  author={Walke, Homer Rich and Black, Kevin and Zhao, Tony Z and Vuong, Quan and Zheng, Chongyi and Hansen-Estruch, Philippe and He, Andre Wang and Myers, Vivek and Kim, Moo Jin and Du, Max and others},
  booktitle={Proceedings of the Conference on Robot Learning},
  pages={1723--1736},
  year={2023}
}

@article{wang2024q,
  title={Q-vlm: Post-training quantization for large vision-language models},
  author={Wang, Changyuan and Wang, Ziwei and Xu, Xiuwei and Tang, Yansong and Zhou, Jie and Lu, Jiwen},
  journal={Advances in Neural Information Processing Systems},
  pages={114553--114573},
  year={2024}
}

@inproceedings{xiao2023smoothquant,
  title={Smoothquant: Accurate and efficient post-training quantization for large language models},
  author={Xiao, Guangxuan and Lin, Ji and Seznec, Mickael and Wu, Hao and Demouth, Julien and Han, Song},
  booktitle={Proceedings of the International Conference on Machine Learning},
  pages={38087--38099},
  year={2023}
}

@article{xu2026qvla,
  title={QVLA: Not All Channels Are Equal in Vision-Language-Action Model's Quantization},
  author={Xu, Yuhao and Yang, Yantai and Fan, Zhenyang and Liu, Yufan and Li, Yuming and Li, Bing and Zhang, Zhipeng},
  journal={arXiv preprint arXiv:2602.03782},
  year={2026}
}

@article{yin2025deepthinkvla,
  title={Deepthinkvla: Enhancing reasoning capability of vision-language-action models},
  author={Yin, Cheng and Lin, Yankai and Xu, Wang and Tam, Sikyuen and Zeng, Xiangrui and Liu, Zhiyuan and Yin, Zhouping},
  journal={arXiv preprint arXiv:2511.15669},
  year={2025}
}

@inproceedings{zhang2026quantvla,
  title={QuantVLA: Scale-Calibrated Post-Training Quantization for Vision-Language-Action Models},
  author={Zhang, Jingxuan and Hsieh, Yunta and Wang, Zhongwei and Lin, Haokun and Wang, Xin and Wang, Ziqi and Lei, Yingtie and Zhang, Mi},
  booktitle={Proceedings of the IEEE Conference on Computer Vision and Pattern Recognition},
  pages={1--11},
  year={2026}
}

@article{zhong2025survey,
  title={A survey on vision-language-action models: An action tokenization perspective},
  author={Zhong, Yifan and Bai, Fengshuo and Cai, Shaofei and Huang, Xuchuan and Chen, Zhang and Zhang, Xiaowei and Wang, Yuanfei and Guo, Shaoyang and Guan, Tianrui and Lui, Ka Nam and others},
  journal={arXiv preprint arXiv:2507.01925},
  year={2025}
}

@inproceedings{zitkovich2023rt,
  title={Rt-2: Vision-language-action models transfer web knowledge to robotic control},
  author={Zitkovich, Brianna and Yu, Tianhe and Xu, Sichun and Xu, Peng and Xiao, Ted and Xia, Fei and Wu, Jialin and Wohlhart, Paul and Welker, Stefan and Wahid, Ayzaan and others},
  booktitle={Proceedings of the Conference on Robot Learning},
  pages={2165--2183},
  year={2023}
}

\clearpage
\appendix

\begin{figure*}[t]
    \centering
    \includegraphics[width=1\textwidth]{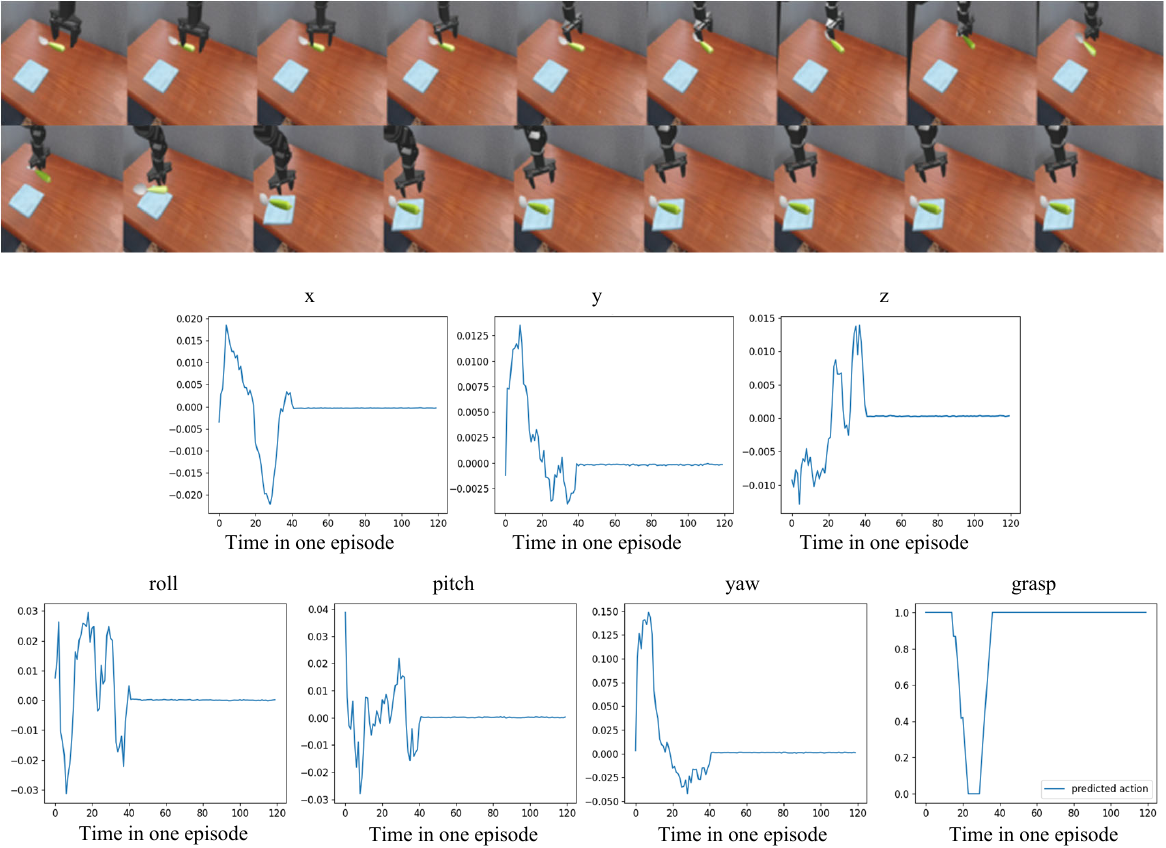}
    \caption{Task: \textit{Put Spoon on Towel} (WidowX Robot). The top panel illustrates the seamless sequential execution by the DA-PTQ quantized model. The bottom panel displays the generated temporal action curves across 7 degrees of freedom, highlighting the absence of quantization-induced oscillations and the smoothness of the low-bit control signals.}
    \label{fig:traj_spoon}
\end{figure*}

\section{Detailed Experimental Setup}
\label{sec:appendix_setup}
To ensure full reproducibility of our DA-PTQ framework, we detail the exact hyperparameters used in our calibration pipeline. Our framework is implemented in PyTorch and deployed on NVIDIA RTX 5090 GPUs, executing entirely post-training with zero fine-tuning overhead. All intermediate calibration activations are cast to BFloat16 to maintain numerical stability. A comprehensive summary of all core hyperparameters is provided in Table~\ref{tab:hyperparameters}.

\begin{table}[b]
    \centering
    \caption{Summary of core hyperparameters used in the DA-PTQ calibration pipeline.}
    \label{tab:hyperparameters}
    \renewcommand{\arraystretch}{1.}
    \setlength{\tabcolsep}{10pt}
    \resizebox{\linewidth}{!}{
    \begin{tabular}{llc}
        \toprule
        \textbf{Module} & \textbf{Parameter} & \textbf{Value}  \\
        \midrule
        \multirow{4}{*}{\textit{Calibration}} 
        & Calibration Steps & $512$  \\
        & Batch Size & $1$  \\
        & Spatial Bins & $6$  \\
        & Warmup Steps & $128$  \\
        \midrule
        \multirow{6}{*}{\textit{DA-MPA}}
        & Probe Steps & $16$  \\
        & Damping Factor ($\lambda$) & $3 \times 10^{-4}$  \\
        & Translation Weight ($w_{trans}$) & $1.8$  \\
        & Rotation Weight ($w_{rot}$) & $0.15$  \\
        & Scaling Gain & $1.6$  \\
        & Retention Ratio ($k$) & $30\%$  \\
        \midrule
        \multirow{4}{*}{\textit{CSRC}}
        & SVD Block Size & $16 \times 16$  \\
        & Smoothing Factor ($\lambda_{smooth}$) & $0.15$  \\
        & Group Size & $32$  \\
        & Shrinkage Factor & $0.55$  \\
        \bottomrule
    \end{tabular}}
\end{table}

\vspace{1.5mm}
\noindent\textbf{Calibration Dataset and Statistics Collection.} We construct our calibration set $\mathcal{D}$ using data sampled from the \textbf{Bridge V2} dataset, which provides a rich and representative distribution of diverse tabletop manipulation trajectories. To prevent the calibration statistics from overfitting to a specific spatial region or a narrow subset of trajectories, we enable \textit{spatial balanced calibration}. This partitions the 3D workspace into spatial bins, applying a warmup phase to ensure the accumulated statistical moments ($\boldsymbol{\mu}_l^{\text{FP}}, \boldsymbol{\Sigma}_l^{\text{FP}}$ and later $\boldsymbol{\mu}_l^{\text{Q}}, \boldsymbol{\Sigma}_l^{\text{Q}}$) robustly reflect the global activation distribution.

\begin{figure*}[t]
    \centering
    \includegraphics[width=1\textwidth]{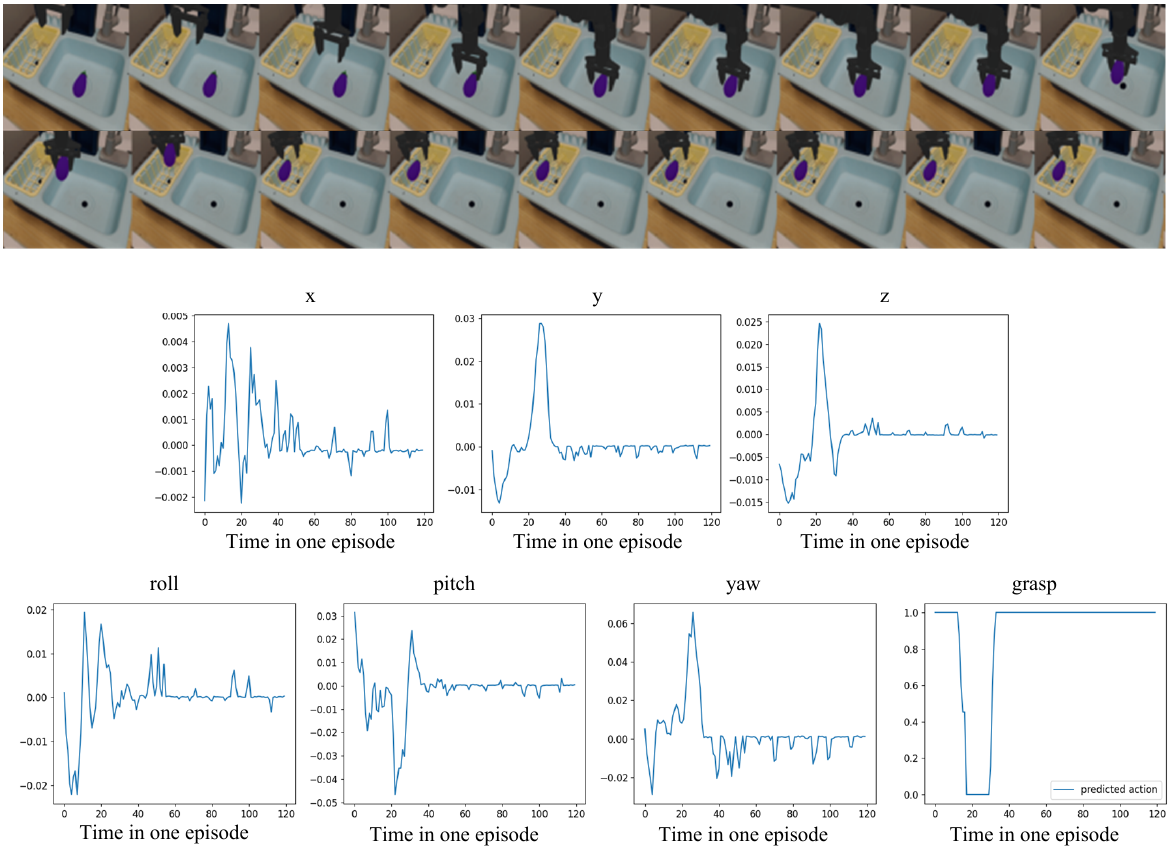}
    \caption{Task: \textit{Put Eggplant in Yellow Basket} (WidowX Robot). DA-PTQ maintains precise and dynamically stable motor commands throughout the episode, guiding the end-effector to successfully complete the manipulation without kinematic drift.}
    \label{fig:traj_eggplant}
\end{figure*}

\begin{figure*}[t]
    \centering
    \includegraphics[width=1\textwidth]{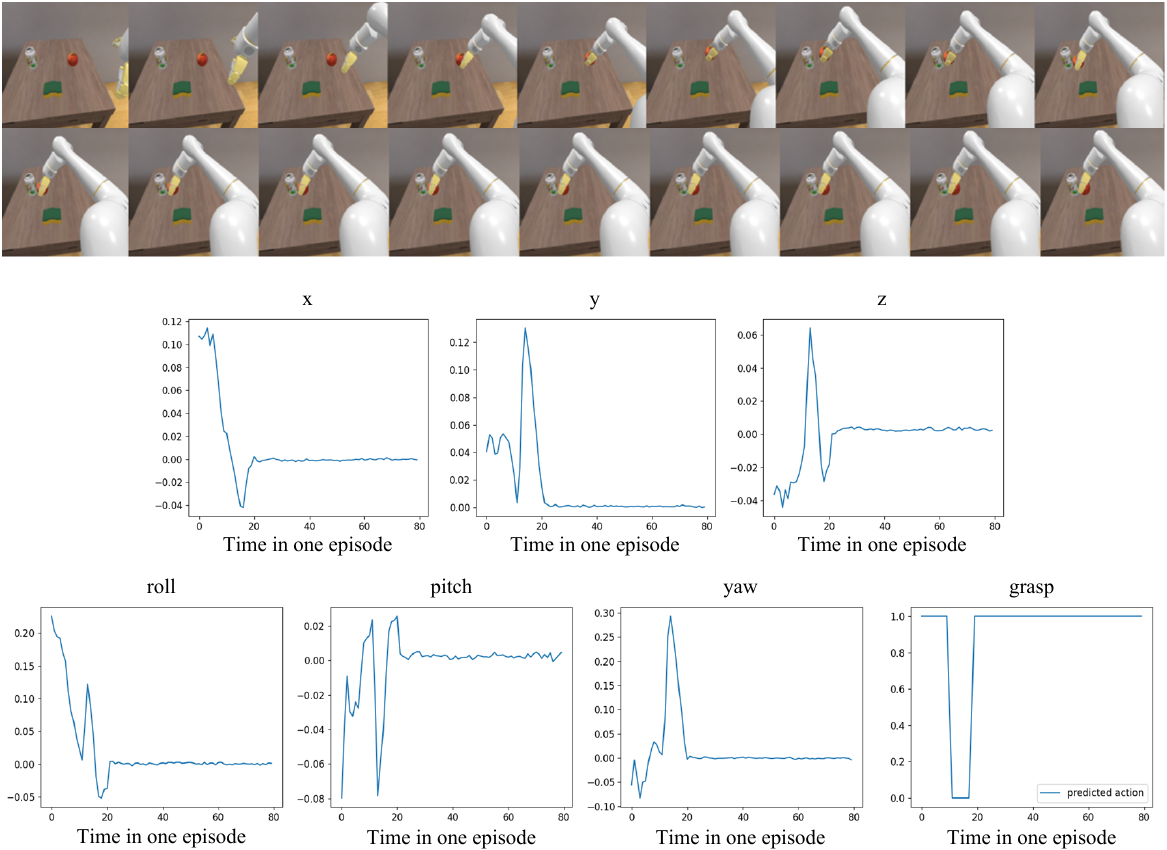}
    \caption{Task: \textit{Move Near} (Google Robot). This qualitative result demonstrates the cross-embodiment generalizability of DA-PTQ. Even deployed on a completely different robotic kinematic structure, the 4-bit framework generates highly accurate and coherent continuous control commands.}
    \label{fig:traj_movenear}
\end{figure*}

\vspace{1.5mm}
\noindent\textbf{Temporal Drift-Aware Allocation (DA-MPA).}
The MPA module specifically targets the MLP blocks within the Diffusion Transformer (DiT) action head, skipping the final $2$ blocks which are naturally preserved due to their proximity to the continuous output layer. The layer-wise drift sensitivity $\phi_l$ is dynamically profiled over consecutive probe steps. 
When calculating the per-dimension drift propagation scores via the structural Jacobian, we apply a Tikhonov damping factor for stable matrix inversion. Recognizing that spatial deviations are highly asymmetric in physical tabletop manipulation—where minor translation errors often lead to catastrophic grasp failures, whereas rotational deviations can be partially absorbed by the mechanical compliance of the gripper—we heavily penalize translational errors.
Based on the profiled sensitivities, we establish a high-precision retention ratio of $k=30\%$. Thus, the top $30\%$ most drift-sensitive DiT layers are safely retained in 16-bit (BF16), while the remaining $70\%$ are aggressively quantized to 4-bit (W4) to maximize inference efficiency.

\vspace{1.5mm}
\noindent\textbf{Cross-Space Representation Compensation (CSRC).} To align the cross-space distributions without introducing any inference latency overhead, we apply our compensation strategy coupled with an outlier pre-rotation mechanism. LLM-based architectures typically suffer from massive activation outliers; thus, the input and output activations are orthogonalized using block-wise rotations computed via SVD. To avoid over-correcting and distorting the underlying feature semantics, we apply a smoothing factor to the singular values. All derived affine compensation matrices ($\mathbf{M}_l$) and bias terms ($\mathbf{d}_l$) are algebraically folded back into the quantized weights prior to real-world deployment.

\section{Qualitative Trajectory Analysis}
\label{sec:appendix_trajectory}
To comprehensively validate the generalizability and control fidelity of DA-PTQ, we visualize the spatial execution and the underlying continuous control commands generated by our quantized policy. We select three distinct, long-horizon manipulation tasks spanning two completely different robotic embodiments: \textit{Put Spoon on Towel} and \textit{Put Eggplant in Yellow Basket} on the WidowX robot, alongside \textit{Move Near} on the Google Robot. The qualitative results are presented in Figures~\ref{fig:traj_spoon}, \ref{fig:traj_eggplant}, and \ref{fig:traj_movenear}.

As shown in the top panels (filmstrips) of these figures, the DA-PTQ quantized policy successfully completes the tasks with fluid, precise, and human-like movements. Despite operating predominantly at an aggressive 4-bit precision (W4), the robotic end-effector exhibits no observable hesitation, kinematic drift, or erratic grasping behaviors. It faithfully handles the objects and navigates complex workspace configurations, demonstrating that DA-PTQ effectively neutralizes the geometric amplification of quantization noise that typically plagues low-bit embodied control.

To provide a granular view of this kinematic stability, the bottom panels detail the step-by-step action predictions across all 7 degrees of freedom (3D translation, 3D rotation, and gripper state) during the entire episode. A hallmark of severe quantization degradation in continuous control is the emergence of high-variance, high-frequency oscillations in the predicted actions. However, as evident in the temporal plots, DA-PTQ ensures that the generated motor commands remain remarkably smooth and dynamically stable. The translational ($x, y, z$) and rotational (roll, pitch, yaw) commands converge smoothly to their target states without overshooting, while the discrete gripper actions are executed decisively. These highly coherent 7-DoF control curves confirm our method's robust capacity to maintain near-lossless, high-fidelity continuous control across diverse physical environments and kinematic structures.

\end{document}